\documentclass{bmvc2k}
\usepackage[table,xcdraw]{xcolor}
\usepackage{amsfonts}
\usepackage{graphicx}
\usepackage{mathtools}
\usepackage{booktabs}
\usepackage{subfigure} 
\usepackage{fix-cm}
\usepackage[scaled=.8]{beramono}
\usepackage{hhline}

\title{Tackling the Unannotated: Scene Graph Generation with Bias-Reduced Models}

\addauthor{Tzu-Jui Julius Wang}{tzu-jui.wang@aalto.fi}{1}
\addauthor{Selen Pehlivan}{selen.pehlivantort@aalto.fi}{1}
\addauthor{Jorma Laaksonen}{jorma.laaksonen@aalto.fi}{1}

\addinstitution{
 Department of Computer Science\\
 School of Science\\
 Aalto University, Espoo, Finland
}

\runninghead{Wang et al}{Tackling the Unannotated: Scene Graph Generation}

\begin{document}

\maketitle

\begin{abstract}
Predicting a scene graph that captures visual entities and their interactions in an image has been considered a crucial step towards full scene comprehension. Recent scene graph generation (SGG) models have shown their capability of capturing the most frequent relations among visual entities. However, the state-of-the-art results are still far from satisfactory, e.g.\@ models can obtain 31\% in \textit{overall recall} R@100, whereas the likewise important \textit{mean class-wise recall} mR@100 is only around 8\% on Visual Genome (VG). The discrepancy between R and mR results urges to shift the focus from pursuing a high R to a high mR with a still competitive R. We suspect that the observed discrepancy stems from both the annotation bias and sparse annotations in VG, in which many visual entity pairs are either not annotated at all or only with a single relation when multiple ones could be valid. To address this particular issue, we propose a novel SGG training scheme that capitalizes on self-learned knowledge. It involves two relation classifiers, one offering a less biased setting for the other to base on. The proposed scheme can be applied to most of the existing SGG models and is straightforward to implement. We observe significant relative improvements in mR (between $+6.6\%$ and $+20.4\%$) and competitive or better R (between $-2.4\%$ and $+0.3\%$) across all standard SGG tasks.
\end{abstract}

\section{Introduction}\label{sec:intro}
Various deep neural network models have been introduced to interpret visual inputs in varying high-level vision tasks, including object detection and segmentation \cite{zhao2019object,long2015fully}, image/video captioning \cite{xu2015show,gao2017video}, and referring expression comprehension \cite{nagaraja2016modeling,yu2018mattnet} etc. While describing visual content and identifying visually grounded entities\footnote{We use \textit{entity} here instead of \textit{object} (as in object detector) to prevent mixing it up with the \textit{object} as in \texttt{subject-object} pair.} from free-form language have been made possible, inferring about the role of entities (either in isolation or relative to other entities) is a crucial step towards full scene comprehension. This fosters the idea of representing an image with a \textit{scene graph} \cite{krishna2017visual} in which the entity classes and their pairwise relations constitute the nodes and directed edges, respectively. In particular, scene graph is the formulation of the visual \texttt{subject-relation-object} triplets (e.g. \texttt{mouse-attached to-laptop}) over entity and relation classes in an image.
The Visual Genome (VG) dataset \cite{krishna2017visual} is introduced to facilitate the advancement in the scene graph generation (SGG) tasks \cite{li2017scene,zellers2018neural,yang2018graph,gu2019scene,chen2019knowledge,tang2019learning,zhang2019graphical,chen2019counterfactual}. However, as first studied in \cite{zellers2018neural}, the annotations in VG bias towards copiously recurring motifs. A strong baseline can be constructed by predicting the relation between entities solely with the priors made on all \texttt{subject-relation-object} triplets from the training set. Moreover, the scene graph of each image is hardly exhaustively annotated with all possible relations \cite{chen2019scene} in which the annotated relations are often less informative \cite{tang2020unbiased} (e.g. labelling a relation as \texttt{girl-near-car} instead of \texttt{girl-behind-car}). Hence, the provided scene graphs inherit the imbalanced and long-tailed natures \cite{liu2019large}. This incurs an apparent complication: the frequent relations, such as \texttt{on}, \texttt{has}, or \texttt{near} can easily be exploited and more often predicted by a model than the rare ones, such as \texttt{walking on}, \texttt{wearing}, or \texttt{in front of}. Then, one can already attain a decent recall rate by simply making predictions from the most frequent relations.

In addition to the imbalanced distribution of the relations, an SGG model can be biased due to the sparse and incomplete annotations \cite{chen2019scene}. Specifically, first, due to the large number of entities in most of the images, the related entity pairs are sparsely identified and annotated, leaving most of the entity pairs presumed as not related. Second, a subject-object pair can often be reasonably related by more than one relation as in \texttt{on}, \texttt{sitting on}, and \texttt{riding} relations for the \texttt{man}-\texttt{bicycle} pair. We refer to the former case as \textbf{annotation False Negative} (aFN), and the latter as \textbf{annotation Partially True Positive} (aPTP). It comes to our attention that nearly all of the SGG models do not address these cases. Overlooking these cases can, (1) wrongly suppress the likelihood of a related entity pair in the predictions, or (2) discourage the model from learning a multi-mode distribution that captures more than one valid relation.

This work aims to address the aFN and aPTP cases in the SGG models. Inspired by knowledge distillation (KD) \cite{44873,zhang2019your,mobahi2020self,tang2020understanding} and co-training framework \cite{qiao2018deep}, we propose a novel SGG framework that regularizes a model's predictions with self-learned knowledge in reasoning the missing information from the annotations. In addition, to alleviate the adverse effect from learning on imbalanced data, the proposed relation models are trained on the hyperspherical space \cite{liu2017sphereface,gidaris2018dynamic}. 
To be more specific, our framework consists of two learners that jointly learn on different assumptions and output spaces. The main learner $\mathcal{F}$ \textit{naively} learns on all relation classes including treating the unannotated entity pairs as the \texttt{no-relation} class. The other \textit{skeptical} learner $\mathcal{G}$ learns on all the relations except \texttt{no-relation}, i.e. no assumption is made for $\mathcal{G}$ on those not annotated being really not related. Over the course of the training, $\mathcal{G}$ transfers the \textit{knowledge} (i.e.\@ the predictions) on both annotated and unannotated entity pairs to $\mathcal{F}$, offering a crucial source of regularization that averts $\mathcal{F}$ learning from the biased annotations (see Section~\ref{sec:method}). 

Our main contributions can be summarized as follows. (1) We propose an SGG framework that addresses biases inherit in the incomplete annotations of the VG dataset. (2) The proposed model is learned on the hyperspherical space, which is shown better at capturing the minority class distribution. (3) The proposed scheme is by design compatible with the most of the existing SGG models and is easily implemented. We demonstrate the effectiveness of the proposed scheme on two state-of-the-art SGGs, Motifs \cite{zellers2018neural} and VCTree \cite{tang2019learning,tang2020unbiased}, and observe significant relative improvements over the baseline counterparts in the mean per-class recall (between $+6.6\%$ and $+20.4\%$) while maintaining competitive overall recall (between $-2.4\%$ to $+0.3\%$) across all the standard SGG tasks \cite{xu2017scene,zellers2018neural} (see Section~\ref{sec:exp}).

\section{Related Work}\label{sec:related_work}
\textbf{Scene Graph Generation.}
SGG offers the possibility of modeling the layout of visual entities and holistically describing the image context. Often being a critical modality in most of the SGG works, contextual modeling is usually implemented by a recurrent neural network (RNN) or a graph neural network (GNN) which learns the underlying structure, e.g. a fully connected \cite{xu2017scene,yang2018graph,chen2019knowledge,chen2019counterfactual}, a chained \cite{zellers2018neural}, or a tree-structured graph \cite{tang2019learning}, among the visual entities. Chen et al. \cite{chen2019counterfactual} formulated SGG as a reinforcement learning problem which aims at directly maximizing the non-differentiable graph-level rewards (e.g. Recall@K). Zhang et al. \cite{zhang2019graphical} proposed a new set of contrastive losses to address (1) repetitive occurrence of entities of the same visual class and (2) confusion caused by the close proximity of similar entities. Despite the improved performance achieved by these state-of-the-art models, SGG models are still facing challenges due to the benchmark datasets being biased towards few majority relation classes. Only recently have some works started to address the imbalance issue, shifting the focus of improvement from the majority relation classes to the minority classes \cite{chen2019knowledge,tang2019learning,tang2020unbiased}. 

In one of the latest SGG works, Tang et al.\@ \cite{tang2020unbiased} proposed a debiased method at the inference phase of SGG models, removing the concentrated probability mass from the majority classes. As one of the few SGG works that address the issues caused by the incomplete annotations, Chen et al. \cite{chen2019scene} proposed a data-efficient scheme that trains the SGG models in a semi-supervised fashion by using a limited labeled and a large amount of unlabeled samples. 

Our proposed method is inspired and closely related to that in \cite{chen2019scene}, but differs in several facets. First, the proposed models co-train two learners, one supervised and another semi-supervised, in one training phase. Second, our supervised learner regularizes itself with the knowledge learned by the semi-supervised learner, significantly improving the predictions on the minority classes. Third, the objective in \cite{chen2019scene} is to train a generative model that generates informative labels on the unannotated pairs given limited amount of labeled data per relation class. However, our objective is to show that an existing SGG model can be greatly improved by harnessing the soft labels on both the annotated and unannotated pairs.\\

\noindent
\textbf{Learning on Imbalanced Data. }
Learning on imbalanced distributions is an inevitable subject to consider in most of the vision tasks \cite{liu2019large,khan2019striking,cui2019class,cao2019learning}. Some common techniques, such as undersampling, oversampling, and class re-weighting \cite{cui2019class} with a cost-sensitive objective \cite{khan2017cost}, have been proven to be effective. The recent studies suggest that a classification model learned with the cross entropy (XE) loss can easily be biased towards majority classes. That is because the XE loss does not explicitly enforce small intra-class variance \cite{gidaris2018dynamic,hayat2019gaussian}. 

A simple modification, which we adopt in this work, is to respectively $L_2$-normalize the input representation and the classifier's weights before calculating the logits (pre-softmax values). This has been shown effective and usually considered as a strong baseline for making predictions in the low-data regime \cite{gidaris2018dynamic,chen2019closer,liu2019large,khan2019striking}. Accordingly, we adopt the same normalization scheme and empirically prove its effectiveness in the SGG problem.\\

\noindent\textbf{Knowledge Distillation. }
Learning a compact "student" model by distilling knowledge from a more complex "teacher" model can lead to a better model than one trained from scratch \cite{44873}. Recently, it has been shown that a neural network model can \textit{self-distill} its knowledge without referring to a teacher model to attain better accuracy \cite{zhang2019your,hahn-choi-2019-self}. \textit{Knowledge distillation} (KD) approaches indicate that even the imperfect predictions (e.g. the predicted class probabilities) can contain useful signals, e.g. the correlations between classes, that help regularize the model to curb the over-confident predictions \cite{tang2020understanding}. We propose two KD variants to combat against the dataset biases (see Section~\ref{sec:intro}) imposed to the models. 



\section{Proposed Framework}\label{sec:method}
    \begin{figure}[t!]
        \centering
        \includegraphics[width=0.95\linewidth,height=0.3\linewidth]{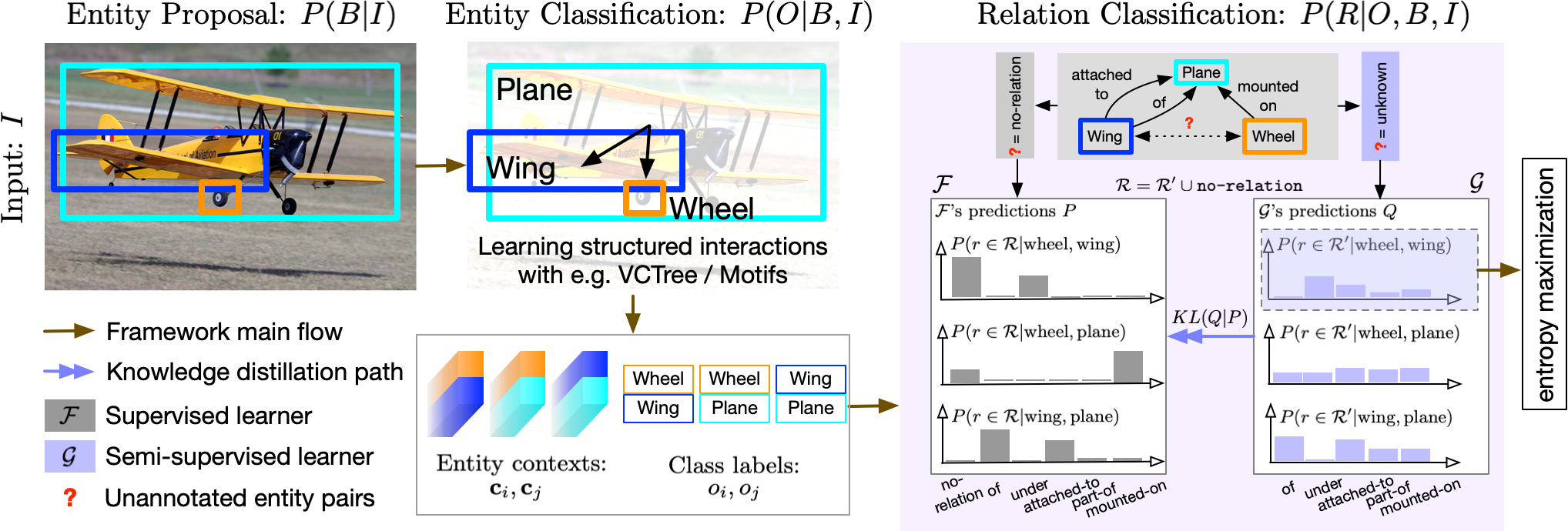}
        \caption{Overview of our proposed two-learner framework. Two learners $\mathcal{F}$ and $\mathcal{G}$ learn on different output spaces $\mathcal{R}$ and $\mathcal{R}'$, respectively,  that make different assumptions on the unannotated entity pairs. Over the course of training, $\mathcal{G}$ transfers its less biased knowledge to $\mathcal{F}$ whose predictions are then regularized to capture more relevant relations among entities. \texttt{Wheel-Wing} refers to the case where the annotation is missing a potential relation \texttt{under}, while \texttt{Wheel-Plane} refers to the case where they could be related by, e.g. \texttt{part of}.}
        \label{fig:overview}
    \end{figure}
A scene graph $\mathcal{S}$ consists of $(B, O, R)$ in which $B=\{B_i | i=1, 2, ..., M\}$, $O=\{o_i \in \mathcal{O} | i=1, 2, ..., M\}$, and $R=\{r_{i,j} \in \mathcal{R} | i,j=1,...,M\}$ are the sets of candidate bounding boxes, entity class labels, and relation class labels,  respectively. In this work, we consider $|\mathcal{O}|=150$ most frequent entity classes and $|\mathcal{R}|=51$ (fifty plus \texttt{no-relation}) most frequent relation classes appearing in the VG dataset. Given an image $I$, one would like to model $P(\mathcal{S}|I)$ which we decompose into three components, i.e. $P(\mathcal{S}|I) = P(R|O, B, I)P(O|B, I)P(B|I)$ as illustrated in Figure~\ref{fig:overview}. The \textbf{entity proposal} component $P(B|I)$ is to generate the bounding box proposals, i.e. the locations and sizes of entities. The \textbf{entity classification} component $P(O|B, I)$ models the entity class probabilities of each proposal in $B$. The \textbf{relation classification} component $P(R|O, B, I)$ is to model the relation class probabilities of every two proposals in $B$ along with their entity classes $O$. Our relation classification component capitalizes on knowledge distillation (KD) and aims at reducing biases in the relation classifiers. We describe the entity proposal and classification components in Section~\ref{sec:method_sub1} and propose our relation classification component and two KD schemes in Section~\ref{sec:method_sub2}. 
\subsection{Entity Proposal and Classification}\label{sec:method_sub1}
\textbf{Entity Proposal Component $P(B|I)$. }We model $P(B|I)$ with Faster R-CNN \cite{ren2015faster} entity detector. We use ResNeXt-101-FPN \cite{xie2017aggregated} (following \cite{tang2020unbiased}) as the backbone network to extract the bounding boxes $B_i$ and their visual representations $\{\textbf{f}_i \in \mathbb{R}^{4096} | i=1,...,M\}$. The same backbone network is used to extract the visual representations $\{\textbf{f}_{i,j} \in \mathbb{R}^{4096} | i,j=1,...,M\}$ from the union region of every two proposals (i.e. the smallest bounding box that covers the two proposals) for the later use (see Section~\ref{sec:method_sub2_2}). As the entity proposal component, we use the model pre-trained on entity proposal labels from VG \cite{ren2015faster} and freeze it when training two following components.\\
\textbf{Entity Classification Component $P(O|B, I)$. } This component comprises a \textit{context module} that takes $\{\textbf{f}_i | i=1,...,M\}$ and outputs the \textit{entity context}\footnote{referred as \textit{object context} in \cite{zellers2018neural,tang2019learning} or contextual cues in \cite{chen2019knowledge}.} $\{\textbf{c}_i\}$ ($\textbf{c}_i \in \mathbb{R}^{512}$) with which $P(O|B, I)$ is modeled. We adopt the context module proposed in either Motifs \cite{zellers2018neural} or VCTree \cite{tang2019learning,tang2020unbiased} (other choices \cite{yang2018graph,chen2019knowledge} are also reasonable). Given $\{\textbf{f}_i\}$, Motifs models $\textbf{c}_i$ with a bi-directional LSTM, while VCTree learns a hierarchical tree structure on which a bi-directional TreeLSTM \cite{tai2015improved} is constructed. The entity contexts capture the structured (e.g.\@ chained or hierarchical) interactions among the visual entities. The classification component is trained on the standard cross-entropy loss (i.e.\@ $L_c$) given the entity class labels. 
\subsection{Relation Classification with Two Learners}\label{sec:method_sub2}
Our proposed relation classification component modeling $P(R|O,B,I)$ consists of two relation learners, a main supervised learner $\mathcal{F}$ and an auxiliary semi-supervised learner $\mathcal{G}$ (see Figure~\ref{fig:overview}). $\mathcal{F}$ learns on both the relation labels and $\mathcal{G}$'s knowledge. This two-learner scheme is introduced to alleviate the biases from the dataset (discussed in Section~\ref{sec:intro}) brought into the models. We elaborate the two learners respectively in the following sections.

\subsubsection{Supervised Learner $\mathcal{F}$}\label{sec:method_sub2_1}
Here we model $P(r \in \mathcal{R} |o_i, o_j)$ for the supervised learner $\mathcal{F}$, in which $o_i,o_j$ represent the entity classes for the $i^{\text{th}}$ and $j^{\text{th}}$ proposals. We first separately learn the representations for subjects $o_i$ and objects $o_j$ as $\textbf{c}_i^\mathcal{F} = W_s^{\mathcal{F}} \textbf{c}_i$ and $\textbf{c}_j^\mathcal{F} = W_o^{\mathcal{F}} \textbf{c}_j$, respectively with their entity contexts $\textbf{c}_i$ and $\textbf{c}_j$. $W_s^{\mathcal{F}}, W_o^{\mathcal{F}} \in \mathbb{R}^{512 \times 512}$ are the parameter matrices for subjects and objects, and $W^{\mathcal{F}}$ is that for the relations. Now, the relation class probabilities are modeled as
    \begin{align}
        P(r \in \mathcal{R}| o_i, o_j) = \textbf{p}_{i,j} &= \text{softmax} \big(
            \gamma \cdot \sigma_{L_2}(W^{\mathcal{F}})
            \sigma_{L_2}(\textbf{g}_{i,j}^{\mathcal{F}}) + \textbf{s}_{o_i,o_j}
        \big ),\quad W^{\mathcal{F}} \in \mathbb{R}^{|\mathcal{R}|\times 4096} \label{eq:f_prob}\\
        \textbf{g}_{i,j}^{\mathcal{F}} &= 
            W_c^{\mathcal{F}}
            [\textbf{c}_i^{\mathcal{F}}, \textbf{c}_j^{\mathcal{F}}] \odot \textbf{f}_{i,j},\quad
                W_c^{\mathcal{F}} \in \mathbb{R}^{4096 \times 1024}
    \end{align}
where $\sigma_{L_2}(\cdot)$ $L_2$-normalizes the vector or each row of the matrix, $[,]$ denotes concatenation of two vectors, and $\odot$ denotes element-wise multiplication. $\textbf{s}_{o_i,o_j}$ is the (constant) frequency prior of $P(r|o_i,o_j)$ pre-calculated from the training set \cite{zellers2018neural}. $\gamma$ is the radius of the hyperspherical space \cite{liu2017sphereface}. Later in Section~\ref{sec:exp_stoa_cpr}, we will show that $L_2$-normalization of both $W^{\mathcal{F}}$ and $\textbf{g}_{i,j}^{\mathcal{F}}$ consistently improves the predictions on the minority relation classes.

The per-image loss for $\mathcal{F}$ is then given by
    \begin{equation}\label{eq:f_nll}
        L_{\mathcal{F}} = -\frac{1}{|R|}
            \sum_{r_{i,j} \in R}
            \log \textbf{p}_{i,j}[r_{i,j}],
    \end{equation}
where $r_{i,j}$ is the target relation class for $o_i,o_j$, and $\textbf{p}_{i,j}[r_{i,j}] = P(r_{i,j}|o_i,o_j) \in [0,1]$ denotes the $r_{i,j}^{\text{th}}$ element of $\textbf{p}_{i,j}$, i.e. the probability of relation $r_{i,j}$. Learning $\mathcal{F}$ purely with Eq.~($\ref{eq:f_nll}$) may impose undesirable biases into the classifier. $P(r_{i,j}|o_i,o_j)$ is encouraged to be maximized (and $P(r \in \mathcal{R}, r \neq r_{i,j}|o_i,o_j)$ to be suppressed) despite that (1) $(o_i, o_j)$ contains a valid but not annotated relation (the aFN cases) and (2) $r \neq r_{i,j}$ could be valid relations between $o_i,o_j$ (the aPTP cases). We address the biases stemming from these two cases with the other classifier $\mathcal{G}$ along with the proposed two KD schemes next.

\subsubsection{Semi-supervised Learner $\mathcal{G}$ and Knowledge Distillation}\label{sec:method_sub2_2}
\textbf{Semi-supervised Learner $\mathcal{G}$.} We co-train the other classifier $\mathcal{G}$ that learns in a semi-supervised fashion. Instead of learning on $\mathcal{R}$, $\mathcal{G}$ learns on $\mathcal{R}' = \mathcal{R} \backslash \{\texttt{no-relation}\}$ while being encouraged to detect as many relations as possible in $\{\texttt{no-relation}\}$. This is to address the fact that \texttt{no-relation} is an artificial class to accommodate the unannotated entity pairs. Specifically, similar to $\mathcal{F}$, given $\textbf{c}_i,\textbf{c}_j$, $\mathcal{G}$ learns a different set of subject and object representations (for establishing a different perspective from that of $\mathcal{F}$) as $\textbf{c}_i^\mathcal{G} = W_s^{\mathcal{G}} \textbf{c}_i$ and $\textbf{c}_j^\mathcal{G} = W_o^{\mathcal{G}} \textbf{c}_j$, respectively, where $W_s^{\mathcal{G}},\, W_o^{\mathcal{G}} \in \mathbb{R}^{512 \times 512}$. Next, the classifier is given by
    \begin{align}
        P(r \in \mathcal{R}' | o_i, o_j) &= \textbf{q}_{i,j} = \text{softmax} \big(
        \gamma \cdot 
            \sigma_{L_2}(W^{\mathcal{G}})
            \sigma_{L_2}(\textbf{g}_{i,j}^{\mathcal{G}})
        \big ),\quad W^{\mathcal{G}} \in \mathbb{R}^{|\mathcal{R}'| \times 4096} \label{eq:g_prob}\\
        \textbf{g}_{i,j}^{\mathcal{G}} &= 
            W_c^{\mathcal{G}}
            [\textbf{c}_i^{\mathcal{G}}, \textbf{c}_j^{\mathcal{G}}]  \odot \textbf{f}_{i,j}, \quad 
                W_c^{\mathcal{G}} \in \mathbb{R}^{4096 \times 1024}
    \end{align}
Comparing Eqs.~(\ref{eq:f_prob}) and (\ref{eq:g_prob}), one can notice that $\mathcal{G}$ learns without the prior $\textbf{s}_{o_i,o_j}$. We argue that the prior, on the one hand, offers a strong baseline for the classifier to learn from, but on the other hand, can bias the classifier towards the majority classes as $\textbf{s}_{o_i,o_j}$ statistically speaks more for them. In addition, adding the same prior as $\mathcal{F}$ to $\mathcal{G}$ can end up learning two classifiers of similar behavior.

The loss function $L_{\mathcal{G}}$ for $\mathcal{G}$ is defined over disjoint sets (1) $R^L=\{ (i,j) | r_{i,j} \in \mathcal{R}'\}$, the set of the entity pairs annotated with a relation in $\mathcal{R}'$ and (2) $R^U=\{ (i,j) | r_{i,j} \not\in \mathcal{R}'\}$, the set of the unannotated entity pairs, in which $|R^L \cup R^U| = |R|$,
    \begin{align}
        L_{\mathcal{G}} = -\frac{1}{|R^{L}|}
            \sum_{(i,j) \in R^{L}}
            \log \,&\textbf{q}^{L}_{i,j}[r_{i,j}] 
            - \lambda_{\mathcal{G}} \frac{1}{|R^{U}|} 
                \sum_{(i,j) \in R^{U}} \mathcal{H}(\textbf{q}_{i,j}^U), \label{eq:g_nll} 
    \end{align}
where $\textbf{q}^{L}_{i,j}=\textbf{q}_{i,j},\, \forall (i,j) \in R^L$, and $\textbf{q}_{i,j}^U=\textbf{q}_{i,j},\, \forall (i,j) \in R^U$. The role of $\mathcal{H}(\textbf{q}_{i,j}^U)$, the entropy of the probability distribution $\textbf{q}_{i,j}^U$ is to be maximized so that $\mathcal{G}$ is encouraged to detect as many relations as possible. $\lambda_{\mathcal{G}}$ is a hyperparameter.
\vspace{2pt}\\
\textbf{Knowledge Distillation.} $\mathcal{G}$ transfers its knowledge to $\mathcal{F}$ through a knowledge distillation (KD) loss $L_{\mathcal{G} \rightarrow \mathcal{F}}$,
    \begin{equation}\label{eq:kt}
        L_{\mathcal{G} \rightarrow \mathcal{F}} = \lambda_{\mathcal{G} \rightarrow \mathcal{F}} \big( 
         \sum_{(i,j) \in R^L} w_{i,j}^L 
                            KL(\Tilde{\textbf{q}}^{L}_{i,j} || \textbf{p}'^{L}_{i,j}) +
         \sum_{(i,j) \in R^U} w_{i,j}^U 
                            KL(\Tilde{\textbf{q}}_{i,j}^U || 
                            \textbf{p}'^{U}_{i,j})
        \big ),
    \end{equation}
where $\lambda_{\mathcal{G} \rightarrow \mathcal{F}}$ is a hyperparameter and $KL(\cdot | \cdot)$ denotes the Kullback-Leibler divergence between two distributions. $\textbf{p}'^{*}_{i,j}$ denotes the first $|\mathcal{R}'|$ elements of $\textbf{p}^{*}_{i,j}$ calculated via Eq.~(\ref{eq:f_prob}) for $(i,j) \in R^{*}$, where $* \in \{L,U\}$. $\Tilde{\textbf{q}}^{*}_{i,j}$ considers \textit{temperature scaling} before transferring the knowledge \cite{44873}, i.e.
    \begin{align}
        \Tilde{\textbf{q}}^{*}_{i,j} = \text{softmax} \big(
                \gamma \cdot \sigma_{L_2}(W^{\mathcal{G}})
            \sigma_{L_2}(\textbf{g}_{i,j}^{\mathcal{G}}) / T 
                \big ), \, \forall (i,j) \in R^{*}, \quad * \in \{L,U\}, \label{eq:temperature} 
    \end{align}
where $T$ is the temperature hyperparameter. Note that Eq.~(\ref{eq:temperature}) reduces to Eq.~(\ref{eq:g_prob}) when $T=1$. In Eq.~(\ref{eq:kt}), $w_{i,j}^*$ serves as a normalization factor for which we propose two different forms: (1) the \textit{uniform KD} (\texttt{uKD}), i.e. $w_{i,j}^{*} = 1/|R^{*}|$ and (2) the \textit{certainty-based KD} (\texttt{cKD}), i.e. $w_{i,j}^{*} = \sum_{(m,n) \in R^{*}} \mathcal{H}(\textbf{q}^{*}_{m,n}) / \mathcal{H}(\textbf{q}^{*}_{i,j}).$
One can see that a more certain $\textbf{q}^{*}_{i,j}$ (smaller $\mathcal{H}(\textbf{q}^{*}_{i,j})$) yields a larger $w_{i,j}^{*}$, i.e. $\mathcal{F}$ is forced to absorb more knowledge from the relatively confident predictions of $\mathcal{G}$. 
We do not propagate gradients from $\mathcal{G}$ to $\mathcal{F}$ to avert $\mathcal{F}$ affecting $\mathcal{G}$ because, otherwise, one would end up obtaining two classifiers that make similar predictions.

The combined loss $L$ from entity classification (see Section~\ref{sec:method_sub1}) and the three relation classification losses is then $L=L_c + L_{\mathcal{F}} + L_{\mathcal{G}} + L_{\mathcal{G} \rightarrow \mathcal{F}}$.
\section{Experiments}\label{sec:exp}
\subsection{Dataset, Tasks, Metrics}\label{sec:exp_sub1}
\textbf{Dataset.} We evaluate the SGG models on the VG dataset \cite{krishna2017visual}.
We follow the same training, validation and test splits as in \cite{zellers2018neural,chen2019knowledge,chen2019counterfactual,tang2019learning,tang2020unbiased}, where only the most frequent $|\mathcal{O}|=150$ entity classes and $|\mathcal{R}|=50$ relation classes are considered.\\
\noindent\textbf{Tasks.} The SGG tasks, from the easiest to the hardest setups, are (1) \textit{predicate classification} (PrdCls), (2) \textit{scene graph classification} (SGCls), and (3) \textit{scene graph detection} (SGDet) \cite{xu2017scene}. In the PrdCls task, the entity proposals and class labels are provided to the model for predicting the relations among entity pairs. In the SGCls task, only the entity proposals are given and models are used to predict both the entity classes and relations. In the SGDet task, models are used to generate entity proposals and predict their classes and relations. 
Among all three tasks, the models are evaluated either \textit{with} or \textit{without the graph constraint} \cite{zellers2018neural,chen2019knowledge,chen2019counterfactual,zhang2019graphical}. Models with the constraint predict a single relation with the highest possibility for each entity pair. Those without the constraint are allowed to predict multiple relations for each entity pair.\\ 
\noindent\textbf{Metrics. }For the tasks with the constraint, we consider recall@K (R@K) and the mean class-wise recall@K (mR@K), where $K=\{20, 50, 100\}$. For those without the constraint, we consider R@\{50,100\} and mR@\{50,100\}. mR@K has been emphasized more in the most recent SGG works \cite{chen2019counterfactual,tang2019learning,tang2020unbiased} to address the imbalanced relation class distributions. R@K accounts for the proportion of the top K confident predicted relation triplets that are in the ground-truth relation triplets. mR@K calculates the average of all R@Ks, each of which is computed separately for the triplets consisting of each relation class.
Other metrics, e.g. precision@K, false positive and negative rates etc., are not adopted and neither in other SGG works  \cite{zellers2018neural,chen2019knowledge,chen2019counterfactual,zhang2019graphical} because they are considered less fair in evaluating the methods performed on the sparsely annotated dataset such as VG.

\subsection{Training Settings}\label{sec:exp_sub2}
We separately train different models for SGDet, SGCls, and PrdCls on top of the pre-trained Faster R-CNN with ResNeXt-101-FPN backbone (provided by the authors in \cite{tang2020unbiased}). The SGG models is subject to the performance of entity detector, therefore, we choose a more complex backbone network to align with one of the most recent SGG works \cite{tang2020unbiased}, however, a direct comparison to other works \cite{zellers2018neural,tang2019learning,chen2019knowledge,chen2019counterfactual} seems difficult as they use an earlier VGG16 backbone. Motifs \cite{zellers2018neural} and VCTree \cite{tang2019learning} are the two baselines for which we adapt our proposed relation component.\\
\textbf{Hyperparameters. } The radius hyperparameter $\gamma=12$ in Eqs.~(\ref{eq:f_prob},\ref{eq:g_prob},\ref{eq:temperature}), temperature $T=1.5$ in Eq.~(\ref{eq:temperature}), $\lambda_{\mathcal{G}}=0.1$ in Eq.~(\ref{eq:g_nll}), and $\lambda_{\mathcal{G} \rightarrow \mathcal{F}}=0.1$ in Eq.~(\ref{eq:kt}) are selected according to the evaluations on the validation set. \\
\textbf{Training Details. }All models are trained up to 50,000 batch iterations of 12 images using SGD with 0.9 momentum. KD starts only after 10,000 iterations ($\lambda_{\mathcal{G} \rightarrow \mathcal{F}}=0$ from 0--10,000 iterations). Learning rate starts from 0.01 and is linearly increased every batch iteration up to 0.12 until 500 iterations (i.e. the warm-up period). After the warm-up, the learning rate is decayed by 0.2 (0.1) for Motifs (VCTree) models if no improvement in R@100 is observed within two successive validation rounds (2,000 iterations / round). The training is early terminated once the learning rate is decayed for three times.

\subsection{Experimental Results and Comparison with the State of the Art}\label{sec:exp_stoa_cpr}
\begin{figure}[ht!]
        \centering
        \includegraphics[width=0.8\linewidth]{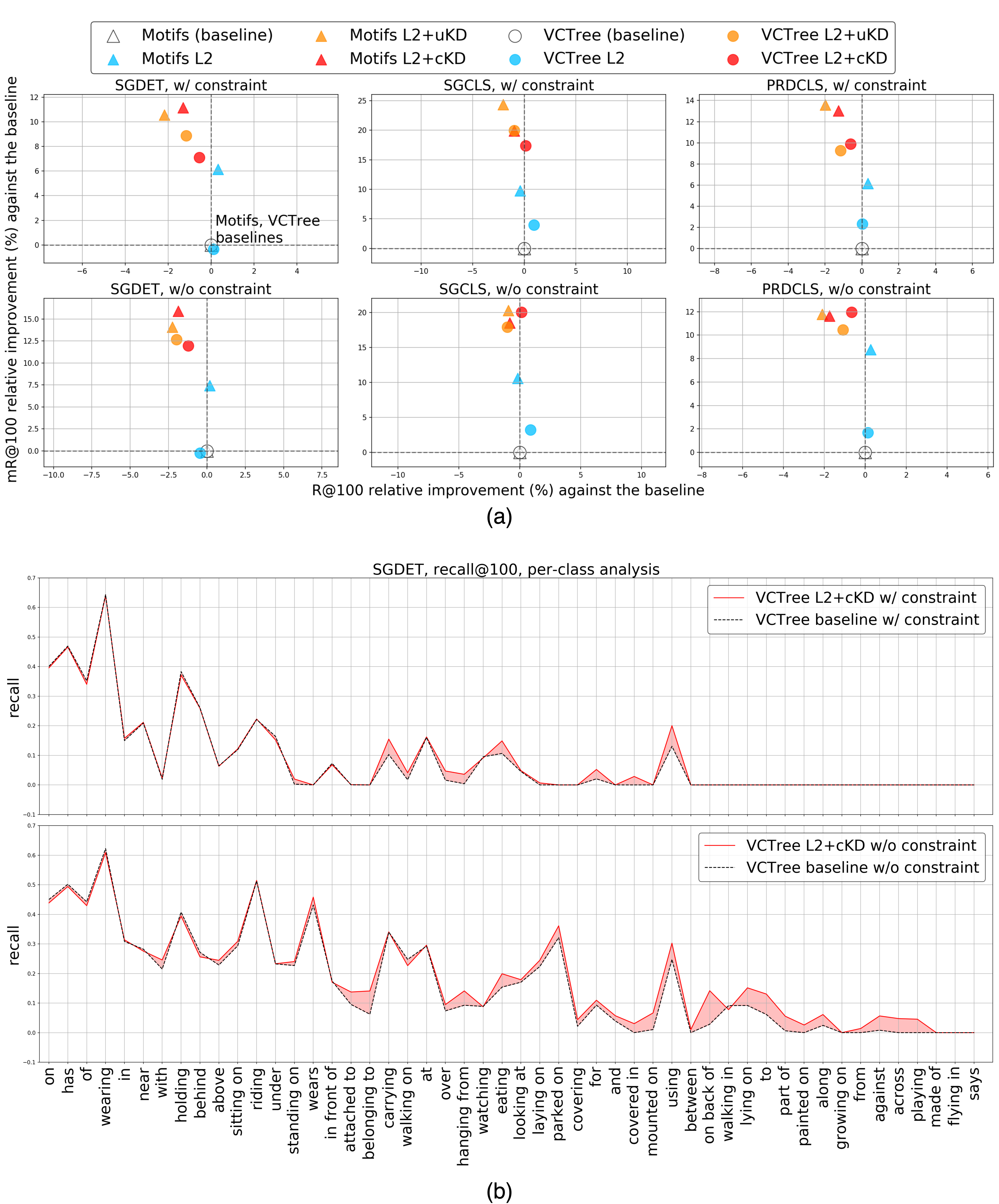}
        \caption{(a) Relative improvements (\%) of models against their baselines, Motifs \cite{zellers2018neural} and VCTree \cite{tang2019learning} in R@100 and mR@100. Two baselines, black hollow $\bigtriangleup$ and $\bigcirc$, serve as reference points for two groups of experiments, respectively. (b) Class-wise recall comparison of \texttt{VCTree baseline} and \texttt{L2+cKD} shows clear improvement in the minority classes.}
        \label{fig:improv}
    \end{figure}

 We conduct extensive quantitative comparisons between our proposed models and the state-of-the-art in Table~\ref{tab:stoa_cpr} and Figure~\ref{fig:improv} to see if the proposed scheme can be used to improve the existing SGG models. \texttt{Motifs/VCTree baseline} and \texttt{L2} differ only in that \texttt{baseline} (which is the re-implementation of the original Motifs/VCTree provided in \cite{tang2020unbiased}) does not $L_2$-normalize the row vectors in $W^{\mathcal{F}}$ and the input features $\textbf{g}^{\mathcal{F}}_{i,j}$ in Eq.~(\ref{eq:f_prob}) while the latter does. Neither \texttt{baseline} nor \texttt{L2} involves learning $\mathcal{G}$. \texttt{L2+uKD} and \texttt{L2+cKD} adopt uniform and certainty-based KD schemes, respectively (see Section~\ref{sec:method_sub2_2}). We inspect the relative improvement ($\%$), i.e. $(\text{R}_A-\text{R}_B)/\text{R}_B \times 100\%$, when comparing the recall $\text{R}_A$ of a proposed model $A$ to $\text{R}_B$ of a baseline model $B$, and similarly for mean class-wise recall mR.
\begin{table}[t!]
\setlength{\tabcolsep}{3.25pt}
\centering
\tiny
\begin{tabular}{|lllllllllllllllllll}
\hline
\multicolumn{1}{|l|}{\textbf{w/ constraint}} & \multicolumn{6}{l|}{\cellcolor[HTML]{FFFFFF}SGDet} & \multicolumn{6}{l|}{\cellcolor[HTML]{FFFFFF}SGCls} & \multicolumn{6}{l|}{\cellcolor[HTML]{FFFFFF}PrdCls} \\
\multicolumn{1}{|l|}{Model} & \multicolumn{3}{l}{\cellcolor[HTML]{FFFFFF}R@\{20,50,100\}} & \multicolumn{3}{l|}{\cellcolor[HTML]{FFFFFF}mR@\{20,50,100\}} & \multicolumn{3}{l}{\cellcolor[HTML]{FFFFFF}R@\{20,50,100\}} & \multicolumn{3}{l|}{\cellcolor[HTML]{FFFFFF}mR@\{20,50,100\}} & \multicolumn{3}{l}{\cellcolor[HTML]{FFFFFF}R@\{20,50,100\}} & \multicolumn{3}{l|}{\cellcolor[HTML]{FFFFFF}mR@\{20,50,100\}} \\ \hline\hline
\multicolumn{1}{|l|}{\texttt{Motifs}$\dagger$ \cite{zellers2018neural}} & 21.4 & 27.2 & \multicolumn{1}{l|}{30.3} & 4.2 & 5.7 & \multicolumn{1}{l|}{6.6} & 32.9 & 35.8 & \multicolumn{1}{l|}{36.5} & 6.3 & 7.7 & \multicolumn{1}{l|}{8.2} & 58.5 & 65.2 & \multicolumn{1}{l|}{67.1} & 10.8 & 14.0 & \multicolumn{1}{l|}{15.3} \\
\multicolumn{1}{|l|}{\texttt{VCTree}$\dagger$ \cite{tang2019learning} } & 22.0 & \textbf{27.9} & \multicolumn{1}{l|}{\textbf{31.3}} & \textbf{5.2} & \textbf{6.9} & \multicolumn{1}{l|}{\textbf{8.0}} & 35.2 & 38.1 & \multicolumn{1}{l|}{38.8} & \textbf{8.2} & \textbf{10.1} & \multicolumn{1}{l|}{\textbf{10.8}} & 60.1 & \textbf{66.4} & \multicolumn{1}{l|}{\textbf{68.1}} & \textbf{14.0} & \textbf{17.9} & \multicolumn{1}{l|}{\textbf{19.4}} \\
\multicolumn{1}{|l|}{\texttt{Routing}$\dagger$ \cite{chen2019knowledge}  } & - & 27.1 & \multicolumn{1}{l|}{29.8} & - & 6.4 & \multicolumn{1}{l|}{7.3} & - & 36.7 & \multicolumn{1}{l|}{37.4} & - & 9.4 & \multicolumn{1}{l|}{10.0} & - & 65.8 & \multicolumn{1}{l|}{67.6} & - & 17.7 & \multicolumn{1}{l|}{19.2} \\
\multicolumn{1}{|l|}{\texttt{Critic}$\dagger$ \cite{chen2019counterfactual} } & \textbf{22.1} & \textbf{27.9} & \multicolumn{1}{l|}{31.2} & - & - & \multicolumn{1}{l|}{-} & \textbf{35.9} & \textbf{39.0} & \multicolumn{1}{l|}{\textbf{39.8}} & - & - & \multicolumn{1}{l|}{-} & \textbf{60.2} & \textbf{66.4} & \multicolumn{1}{l|}{\textbf{68.1}} & - & - & \multicolumn{1}{l|}{-} \\\hline
\multicolumn{1}{|l|}{\texttt{Motifs}$\star$ \cite{tang2020unbiased}} & 25.48 & 32.78 & \multicolumn{1}{l|}{37.16} & 4.98 & 6.75 & \multicolumn{1}{l|}{7.90} & 35.63 & 38.92 & \multicolumn{1}{l|}{39.77} & 6.68 & 8.28 & \multicolumn{1}{l|}{8.81} & 58.46 & 65.18 & \multicolumn{1}{l|}{67.01} & 11.67 & 14.79 & \multicolumn{1}{l|}{16.08} \\
\multicolumn{1}{|l|}{\texttt{Motifs baseline}} & 25.78 & 33.08 & \multicolumn{1}{l|}{37.57} & 5.37 & 7.36 & \multicolumn{1}{l|}{8.62} & \textbf{36.13} & \textbf{39.35} & \multicolumn{1}{l|}{\textbf{40.14}} & 7.32 & 8.94 & \multicolumn{1}{l|}{9.50} & 59.15 & 65.57 & \multicolumn{1}{l|}{67.29} & 13.01 & 16.63 & \multicolumn{1}{l|}{17.89} \\
\multicolumn{1}{|l|}{\texttt{Motifs L2}} & \textbf{25.86} & \textbf{33.22} & \multicolumn{1}{l|}{\textbf{37.69}} & 5.68 & 7.77 & \multicolumn{1}{l|}{9.15} & 36.06 & 39.21 & \multicolumn{1}{l|}{39.98} & 8.01 & 9.80 & \multicolumn{1}{l|}{10.43} & \textbf{59.33} & \textbf{65.76} & \multicolumn{1}{l|}{\textbf{67.51}} & 13.78 & 17.55 & \multicolumn{1}{l|}{18.99} \\
\multicolumn{1}{|l|}{\texttt{Motifs L2+uKD}} & 24.80 & 32.24 & \multicolumn{1}{l|}{36.75} & 5.68 & 7.88 & \multicolumn{1}{l|}{9.53} & 35.11 & 38.50 & \multicolumn{1}{l|}{39.32} & 8.57 & \textbf{10.90} & \multicolumn{1}{l|}{\textbf{11.81}} & 57.37 & 64.14 & \multicolumn{1}{l|}{65.95} & 14.19 & \textbf{18.59} & \multicolumn{1}{l|}{\textbf{20.32}} \\
\multicolumn{1}{|l|}{\texttt{Motifs L2+cKD}} & 25.20 & 32.50 & \multicolumn{1}{l|}{37.08} & \textbf{5.81} & \textbf{8.05} & \multicolumn{1}{l|}{\textbf{9.58}} & 35.58 & 38.93 & \multicolumn{1}{l|}{39.75} & \textbf{8.66} & 10.71 & \multicolumn{1}{l|}{11.39} & 57.73 & 64.58 & \multicolumn{1}{l|}{66.44} & \textbf{14.36} & 18.49 & \multicolumn{1}{l|}{20.22} \\\hline
\multicolumn{1}{|l|}{\texttt{VCTree}$\star$ \cite{tang2020unbiased}} & 24.53 & 31.93 & \multicolumn{1}{l|}{36.21} & 5.38 & 7.44 & \multicolumn{1}{l|}{8.66} & 42.77 & 46.67 & \multicolumn{1}{l|}{47.64} & 9.59 & 11.81 & \multicolumn{1}{l|}{12.52} & 59.02 & 65.42 & \multicolumn{1}{l|}{67.18} & 13.12 & 16.74 & \multicolumn{1}{l|}{18.16} \\
\multicolumn{1}{|l|}{\texttt{VCTree baseline}} & \textbf{25.15} & \textbf{32.23} & \multicolumn{1}{l|}{36.32} & 5.31 & 7.29 & \multicolumn{1}{l|}{8.46} & 41.48 & 45.10 & \multicolumn{1}{l|}{46.04} & 8.54 & 10.48 & \multicolumn{1}{l|}{11.17} & 59.80 & 65.87 & \multicolumn{1}{l|}{67.49} & 13.33 & 16.89 & \multicolumn{1}{l|}{18.23} \\
\multicolumn{1}{|l|}{\texttt{VCTree L2}} & 25.13 & 32.20 & \multicolumn{1}{l|}{\textbf{36.36}} & 5.29 & 7.22 & \multicolumn{1}{l|}{8.43} & \textbf{41.86} & \textbf{45.54} & \multicolumn{1}{l|}{\textbf{46.48}} & 8.85 & 10.90 & \multicolumn{1}{l|}{11.61} & \textbf{59.76} & \textbf{65.90} & \multicolumn{1}{l|}{67.51} & 13.68 & 17.25 & \multicolumn{1}{l|}{18.65} \\
\multicolumn{1}{|l|}{\texttt{VCTree L2+uKD}} & 24.44 & 31.60 & \multicolumn{1}{l|}{35.90} & \textbf{5.68} & 7.69 & \multicolumn{1}{l|}{\textbf{9.21}} & 40.93 & 44.66 & \multicolumn{1}{l|}{45.60} & \textbf{9.90} & \textbf{12.43} & \multicolumn{1}{l|}{\textbf{13.40}} & 58.49 & 65.01 & \multicolumn{1}{l|}{66.71} & 14.20 & 18.24 & \multicolumn{1}{l|}{19.92} \\
\multicolumn{1}{|l|}{\texttt{VCTree L2+cKD}} & 24.84 & 32.02 & \multicolumn{1}{l|}{36.12} & 5.66 & \textbf{7.73} & \multicolumn{1}{l|}{9.06} & 41.41 & 45.17 & \multicolumn{1}{l|}{46.11} & 9.65 & 12.14 & \multicolumn{1}{l|}{13.11} & 59.04 & 65.42 & \multicolumn{1}{l|}{67.07} & \textbf{14.41} & \textbf{18.42} & \multicolumn{1}{l|}{\textbf{20.03}} \\ \hline
\end{tabular}
\begin{tabular}{|lllllllllllll|}
\hline
\multicolumn{1}{|l|}{\textbf{w/o constraint}} & \multicolumn{4}{l|}{\cellcolor[HTML]{FFFFFF}SGDet} & \multicolumn{4}{l|}{\cellcolor[HTML]{FFFFFF}SGCls} & \multicolumn{4}{l|}{\cellcolor[HTML]{FFFFFF}PrdCls} \\
\multicolumn{1}{|l|}{Model} & \multicolumn{2}{l}{\cellcolor[HTML]{FFFFFF}R@\{50,100\}} & \multicolumn{2}{l|}{\cellcolor[HTML]{FFFFFF}mR@\{50,100\}} & \multicolumn{2}{l}{\cellcolor[HTML]{FFFFFF}R@\{50,100\}} & \multicolumn{2}{l|}{\cellcolor[HTML]{FFFFFF}mR@\{50,100\}} & \multicolumn{2}{l}{\cellcolor[HTML]{FFFFFF}R@\{50,100\}} & \multicolumn{2}{l|}{\cellcolor[HTML]{FFFFFF}mR@\{50,100\}}  \\ 
\hline\hline
\multicolumn{1}{|l|}{\texttt{Motifs}$\dagger$ \cite{zellers2018neural}} & 30.5 & \multicolumn{1}{l|}{35.8} & - & \multicolumn{1}{l|}{-} & 44.5 & \multicolumn{1}{l|}{47.7} & - & \multicolumn{1}{l|}{-} & 81.1 & \multicolumn{1}{l|}{88.3} & - & \multicolumn{1}{l|}{-}  \\
\multicolumn{1}{|l|}{\texttt{VCTree}$\dagger$ \cite{tang2019learning} } & - & \multicolumn{1}{l|}{-} & - & \multicolumn{1}{l|}{-} & - & \multicolumn{1}{l|}{-} & - & \multicolumn{1}{l|}{-} & - & \multicolumn{1}{l|}{-} & - & \multicolumn{1}{l|}{-} \\
\multicolumn{1}{|l|}{\texttt{Routing}$\dagger$ \cite{chen2019knowledge} } & 30.9 & \multicolumn{1}{l|}{35.8} & 11.7 & \multicolumn{1}{l|}{16.0} & 45.9 & \multicolumn{1}{l|}{49.0} & 19.8 & \multicolumn{1}{l|}{26.20} & 81.9 & \multicolumn{1}{l|}{88.9} & 36.3 & \multicolumn{1}{l|}{49.0} \\
\multicolumn{1}{|l|}{\texttt{Critic}$\dagger$ \cite{chen2019counterfactual} } & \textbf{31.6} & \multicolumn{1}{l|}{\textbf{36.8}} & - & \multicolumn{1}{l|}{-} & \textbf{48.6} & \multicolumn{1}{l|}{\textbf{52.0}} & - & \multicolumn{1}{l|}{-} & \textbf{83.2} & \multicolumn{1}{l|}{\textbf{90.1}} & - & \multicolumn{1}{l|}{-} \\\cline{1-13}
\multicolumn{1}{|l|}{\texttt{Motifs}$\star$ \cite{tang2020unbiased}} & 36.58 & \multicolumn{1}{l|}{43.43} & - & \multicolumn{1}{l|}{-} & 48.48 & \multicolumn{1}{l|}{51.98} & - & \multicolumn{1}{l|}{-} & 81.02 & \multicolumn{1}{l|}{88.24} & - & \multicolumn{1}{l|}{-} \\
\multicolumn{1}{|l|}{\texttt{Motifs baseline}} & 37.12 & \multicolumn{1}{l|}{43.97} & 12.63 & \multicolumn{1}{l|}{17.07} & \textbf{49.03} & \multicolumn{1}{l|}{\textbf{52.61}} & 18.66 & \multicolumn{1}{l|}{25.01} & 81.95 & \multicolumn{1}{l|}{89.05} & 33.28 & \multicolumn{1}{l|}{45.53} \\
\multicolumn{1}{|l|}{\texttt{Motifs L2}} & \textbf{37.23} & \multicolumn{1}{l|}{\textbf{44.05}} & 13.54 & \multicolumn{1}{l|}{18.34} & 48.97 & \multicolumn{1}{l|}{52.52} & 20.70 & \multicolumn{1}{l|}{27.66} & \textbf{82.22} & \multicolumn{1}{l|}{\textbf{89.29}} & 36.61 & \multicolumn{1}{l|}{49.52}  \\
\multicolumn{1}{|l|}{\texttt{Motifs L2+uKD}} & 35.98 & \multicolumn{1}{l|}{42.98} & 13.95 & \multicolumn{1}{l|}{19.47} & 48.53 & \multicolumn{1}{l|}{52.12} & \textbf{22.73} & \multicolumn{1}{l|}{\textbf{30.09}} & 80.02 & \multicolumn{1}{l|}{87.18} & 36.92 & \multicolumn{1}{l|}{\textbf{50.89}} \\
\multicolumn{1}{|l|}{\texttt{Motifs L2+cKD}} & 36.27 & \multicolumn{1}{l|}{43.15} & \textbf{14.22} & \multicolumn{1}{l|}{\textbf{19.78}} & 48.63 & \multicolumn{1}{l|}{52.17} & 22.07 & \multicolumn{1}{l|}{29.64} & 80.29 & \multicolumn{1}{l|}{87.49} & \textbf{37.15} & \multicolumn{1}{l|}{50.83} \\\cline{1-13}
\multicolumn{1}{|l|}{\texttt{VCTree}$\star$ \cite{tang2020unbiased}} & 35.73 & \multicolumn{1}{l|}{42.34} & - & \multicolumn{1}{l|}{-} & 58.36 & \multicolumn{1}{l|}{62.70} & - & \multicolumn{1}{l|}{-} & 81.63 & \multicolumn{1}{l|}{88.83} & - & \multicolumn{1}{l|}{-}  \\
\multicolumn{1}{|l|}{\texttt{VCTree baseline}} & \textbf{36.25} & \multicolumn{1}{l|}{\textbf{42.90}} & 12.42 & \multicolumn{1}{l|}{16.98} & 56.46 & \multicolumn{1}{l|}{60.59} & 22.26 & \multicolumn{1}{l|}{29.84} & 82.56 & \multicolumn{1}{l|}{89.43} & 34.45 & \multicolumn{1}{l|}{46.82}  \\
\multicolumn{1}{|l|}{\texttt{VCTree L2}} & 36.22 & \multicolumn{1}{l|}{42.71} & 12.53 & \multicolumn{1}{l|}{16.94} & \textbf{56.94} & \multicolumn{1}{l|}{\textbf{61.12}} & 22.82 & \multicolumn{1}{l|}{30.80} & \textbf{82.69} & \multicolumn{1}{l|}{\textbf{89.54}} & 35.12 & \multicolumn{1}{l|}{47.61}  \\
\multicolumn{1}{|l|}{\texttt{VCTree L2+uKD}} & 35.48 & \multicolumn{1}{l|}{42.05} & 13.82 & \multicolumn{1}{l|}{\textbf{19.13}} & 55.97 & \multicolumn{1}{l|}{59.97} & \textbf{26.83} & \multicolumn{1}{l|}{35.19} & 81.43 & \multicolumn{1}{l|}{88.46} & 37.69 & \multicolumn{1}{l|}{51.72}  \\
\multicolumn{1}{|l|}{\texttt{VCTree L2+cKD}} & 35.88 & \multicolumn{1}{l|}{42.38} & \textbf{13.93} & \multicolumn{1}{l|}{19.01} & 56.61 & \multicolumn{1}{l|}{60.67} & 26.80 & \multicolumn{1}{l|}{\textbf{35.82}} & 81.94 & \multicolumn{1}{l|}{88.83} & \textbf{38.35} & \multicolumn{1}{l|}{\textbf{52.42}} \\ \cline{1-13}
\end{tabular}
\vspace{2pt}
\caption{Comparisons with the state of the art. Motifs \cite{zellers2018neural} and VCTree \cite{tang2019learning} baselines are established to be compared. Models with $\dagger$ refer to those with VGG16 as the backbone network while the other moare with ResNeXt-101-FPN. Models with $\star$ refer to the results reported by the authors of \cite{tang2020unbiased}. They are excluded from the comparison due to some modifications to the \texttt{Motifs/VCTree baseline} on which our proposed models are based.}
\label{tab:stoa_cpr}
\vspace{-10pt}
\end{table}
\\
\textbf{\texttt{L2} vs. \texttt{baseline}.} We observe consistent improvements in mR@K across the three SGG tasks in almost all metrics, e.g. $+6.15\%$ in mR@100 with \texttt{Motifs L2} in SGDet with the graph constraint. 
The changes in R@K across all tasks (with and without the constraint) are nearly negligible ($-1\%$ to $1\%$).\\
\textbf{\texttt{L2+cKD} vs. \texttt{L2+uKD}.} \texttt{L2+cKD} performs better than \texttt{L2+uKD} in R@K across all tasks with and without the constraint. However, \texttt{L2+uKD} can sometimes slightly outperforms \texttt{L2+cKD} in mR. For instance, \texttt{VCTree L2+uKD} leads \texttt{L2+cKD} by $0.49\%$ in mR@K but is $1.19\%$ behind in R@K on average in SGDet with the constraint. Although cKD falls short in mR@K in some cases when evaluated with the graph constraint (first row, Figure 2(a)), the difference in mR@K obtained by cKD and uKD usually becomes almost negligible without the graph constraint (second row, Figure 2(a)). Overall, we favor \texttt{L2+cKD} over \texttt{L2+uKD} models here since the former usually produce more balanced results in R@K and mR@K (see Figure~\ref{fig:improv}). Hence, we mainly compare \texttt{L2+cKD} against other models from here on.
\\
\textbf{\texttt{L2+cKD} vs. \texttt{L2}.} \texttt{L2+cKD} significantly improves mR@K at a negligible cost of R@K, e.g. \texttt{VCTree L2+cKD} outperforms \texttt{VCTree L2} by $7.18\%$ in mR@K with only $-0.79\%$ loss in R@K on average in SGDet with the constraint. The relative improvements in mR@K to R@K are even higher in SGCls. This indicates that \texttt{L2+cKD} is more effective in reducing the biases from the imbalanced distributions in VG than \texttt{L2}. \\
\textbf{Improvements in mR@K vs. R@K. }The relative improvements of \texttt{L2+cKD} against \texttt{base\-line} in mR@K ($6.03\%-20.39\%$) are much higher than the losses (at worst, $-1.21\%$) in R@K across all tasks (see Figure~\ref{fig:improv}(a)) with or without the constraint. The main losses in the majority classes are usually negligible compared to the gains in the minority classes (see Figure~\ref{fig:improv}(b)). This indicates that our proposed scheme is more capable of modeling rare relation classes, e.g. \texttt{on back of}, \texttt{lying on}, and \texttt{playing} etc. \\ 
\textbf{Improvements in models w/ constraint vs. w/o constraint.} In SGDet, our proposed models show much more improvements in mR@50 and mR@100 when not under the graph constraint than under it. This suggests that, when not constrained, our models are more able to retrieve more relevant relations (and ignore the less relevant) than \texttt{baseline}. \\
\textbf{Ablation study on temperature $T$ in Eq.~(\ref{eq:temperature}). }From Table~\ref{tab:temp_cpr}, we can see that one can attain a higher mR@K with a larger $T$ in the PrdCls task with \texttt{VCTree L2+cKD}. However, the drop in R@K can become deeper in exchange of lower improvements in mR@K. This can be explained by the fact that the learner $\mathcal{G}$ can provide less informative (i.e. more uniform) predictions with a larger $T$ that overly regularizes $\mathcal{F}$'s predictions.
\begin{table}[ht!]
\setlength{\tabcolsep}{2.5pt}
\centering
\begin{tabular}{|lllllllllll|}
\hline
\rowcolor[HTML]{FFFFFF} 
 & \multicolumn{6}{|l}{\cellcolor[HTML]{FFFFFF}\textbf{with constraint}, PrdCls} & \multicolumn{4}{l|}{\cellcolor[HTML]{FFFFFF}\textbf{w/o constraint}, PrdCls} \\
\rowcolor[HTML]{FFFFFF} 
 & \multicolumn{3}{|l}{\cellcolor[HTML]{FFFFFF}R@\{20,50,100\}} & \multicolumn{3}{l}{\cellcolor[HTML]{FFFFFF}mR@\{20,50,100\}} & \multicolumn{2}{l}{\cellcolor[HTML]{FFFFFF}R@\{50,100\}} & \multicolumn{2}{l|}{\cellcolor[HTML]{FFFFFF}mR@\{50,100\}} \\ \hline
 \hline
$T$=1.0 & \multicolumn{1}{|l}{58.74} & \textbf{65.68} & \multicolumn{1}{l|}{\textbf{67.54}} & 13.66 & 17.75 & \multicolumn{1}{l|}{19.37} & 81.81 & \multicolumn{1}{l|}{\textbf{89.02}} & 36.25 & 49.52 \\
$T$=1.25 & \multicolumn{1}{|l}{58.76} & 65.55 & \multicolumn{1}{l|}{67.39} & 13.6 & 18.15 & \multicolumn{1}{l|}{19.94} & 81.74 & \multicolumn{1}{l|}{88.98} & 37.23 & 51.31 \\
$T$=1.5 & \multicolumn{1}{|l}{\textbf{59.04}} & 65.42 & \multicolumn{1}{l|}{67.07} & \textbf{14.41} & 18.42 & \multicolumn{1}{l|}{20.03} & \textbf{81.94} & \multicolumn{1}{l|}{88.83} & 38.35 & 52.42 \\
$T$=1.75 & \multicolumn{1}{|l}{58.86} & 65.32 & \multicolumn{1}{l|}{66.96} & 14.36 & \textbf{18.5} & \multicolumn{1}{l|}{\textbf{20.18}} & 81.52 & \multicolumn{1}{l|}{88.31} & \textbf{38.88} & \textbf{52.48} \\ \hline
\end{tabular}
\vspace{1pt}
\caption{Ablation study on temperature hyperparameter $T$ in Eq.~(\ref{eq:temperature}) on \texttt{VCTree L2+cKD}.}
\label{tab:temp_cpr}
\end{table}

\section{Conclusion}
We proposed a novel SGG framework consisting of two relation learners in which one learner regularizes the other learner's predictions with the less biased knowledge. This particularly addresses the biases incurred by the incomplete annotations in VG. The quantitative analyses showed that our proposed models are less biased towards the majority relation classes and more capable of capturing more than one relevant relations. Empirically we improved two state-of-the-arts, Motifs \cite{zellers2018neural} and VCTree \cite{tang2019learning} with (and without) the graph constraint, respectively by $11.14\%$ ($14.23\%$) and $7.09\%$ ($12.06\%$) in SGDet, by $19.89\%$ ($18.39\%$) and $17.37\%$ ($20.22\%$) in SGCls, and by $13.02\%$ ($11.64\%$) and $9.87\%$ ($11.96\%$) in PrdCls in mR@100. R@100 results are $-1.3\%$ ($-1.86\%$) at worst across all tasks with (without) the constraint. We believe that the improvements come from the fact that our proposed models alleviate the impact of the biases from the partially annotated and unannotated samples in the popular VG dataset. One of the future directions is to investigate the generalization ability of our models on the rarely seen (few-shot \cite{dornadula2019visual}) or unseen (zero-shot \cite{lu2016visual}) \texttt{subject-relation-object} triplets.
\section*{Acknowledgements}
This work has been funded by the Academy of Finland project number 313988 (DeepGraph), and the European Union's Horizon 2020 research and innovation programme under grant agreement No. 780069 (MeMAD). We also thank the Aalto University's Aalto Science IT project and CSC – IT Center for Science Ltd. for providing computer resources and NVIDIA Corporation for donation of GPU for this research.
\newpage
\bibliography{main}
\end{document}